\documentclass{article}
\usepackage{spconf,hyperref}
\usepackage{cite}
\usepackage{amsmath,amssymb,amsfonts}
\usepackage{algorithmic}
\usepackage{graphicx}
\usepackage{textcomp}
\usepackage{xcolor}
\def\BibTeX{{\rm B\kern-.05em{\sc i\kern-.025em b}\kern-.08em
    T\kern-.1667em\lower.7ex\hbox{E}\kern-.125emX}}
\usepackage{acronym}
\usepackage{comment}
\usepackage{multicol}
\usepackage{multirow}
\usepackage{xcolor}
\usepackage{subcaption}
\usepackage{float}
\usepackage{cleveref}
\usepackage{url}
\usepackage{booktabs}
\usepackage{amsthm}
\usepackage{enumitem}

\def\etal{\emph{et al}.}

\title{GRAPHPL: LEVERAGING GNN FOR EFFICIENT AND ROBUST MODALITIES IMPUTATION IN PATCHWORK LEARNING}
\name{Xingjian Hu$^{\star}$, Zuoyu Yan$^{\dagger}$, Jianhua Zhu$^{\star}$, Liangcai Gao$^{\star}$, Fei Wang$^{\dagger}$, Tengfei Ma$^*$}
\address{$^{\star}$ Wangxuan Institute of Computer Technology, Peking University, Beijing\\ 
        E-mail: \{huxingjian, glc\}@pku.edu.cn\\
        $^{\dagger}$ Weill Cornell Medicine, Cornell University, New York\\
        $^*$ Biomedical Informatics, Stony Brook University, New York}

\begin{document}
\acrodef{POE}[POE]{product-of-experts}
\acrodef{MOE}[MOE]{mixture-of-experts}
\acrodef{GNN}[GNNs]{Graph Neural Networks}
\acrodef{EHR}[EHR]{Electronic Health Record}
\definecolor{darkorange}{RGB}{255,140,0}
\definecolor{mplpurple}{RGB}{128, 0, 128}
\definecolor{mplgreen}{RGB}{0, 128, 0}
\definecolor{mplblue}{RGB}{0, 0, 255}
\definecolor{darkorange}{RGB}{255,140,0}
\definecolor{mplred}{RGB}{255, 0, 0}
\definecolor{train}{RGB}{177,217,167}
\definecolor{infer}{RGB}{255,165,114}
%
\maketitle
\begin{abstract}
Current research on distributed multi-modal learning typically assumes that clients can access complete information across all modalities, which may not hold in practice. In this paper, we explore patchwork learning, in which the modalities available to different clients vary, and the objective is to impute the missing modalities for each client in an unsupervised manner. 
Existing methods are shown not to fully utilize the modality information as they tend to rely on only a subset of the observed modalities. 
To address this issue, we propose GraphPL, which combines graph neural networks with patchwork learning to flexibly integrate all observed modalities and remains robust with noisy inputs.
Experimental results show that GraphPL achieves SOTA performance on benchmark datasets. Our results on real-world distributed electronic health record dataset show GraphPL learns strong downstream features and enables tasks like disease prediction via superior modality imputation.
\end{abstract}
\section{Introduction}
\label{sec:intro}

\begin{figure}[t]
  \centering
   \includegraphics[width=\linewidth]{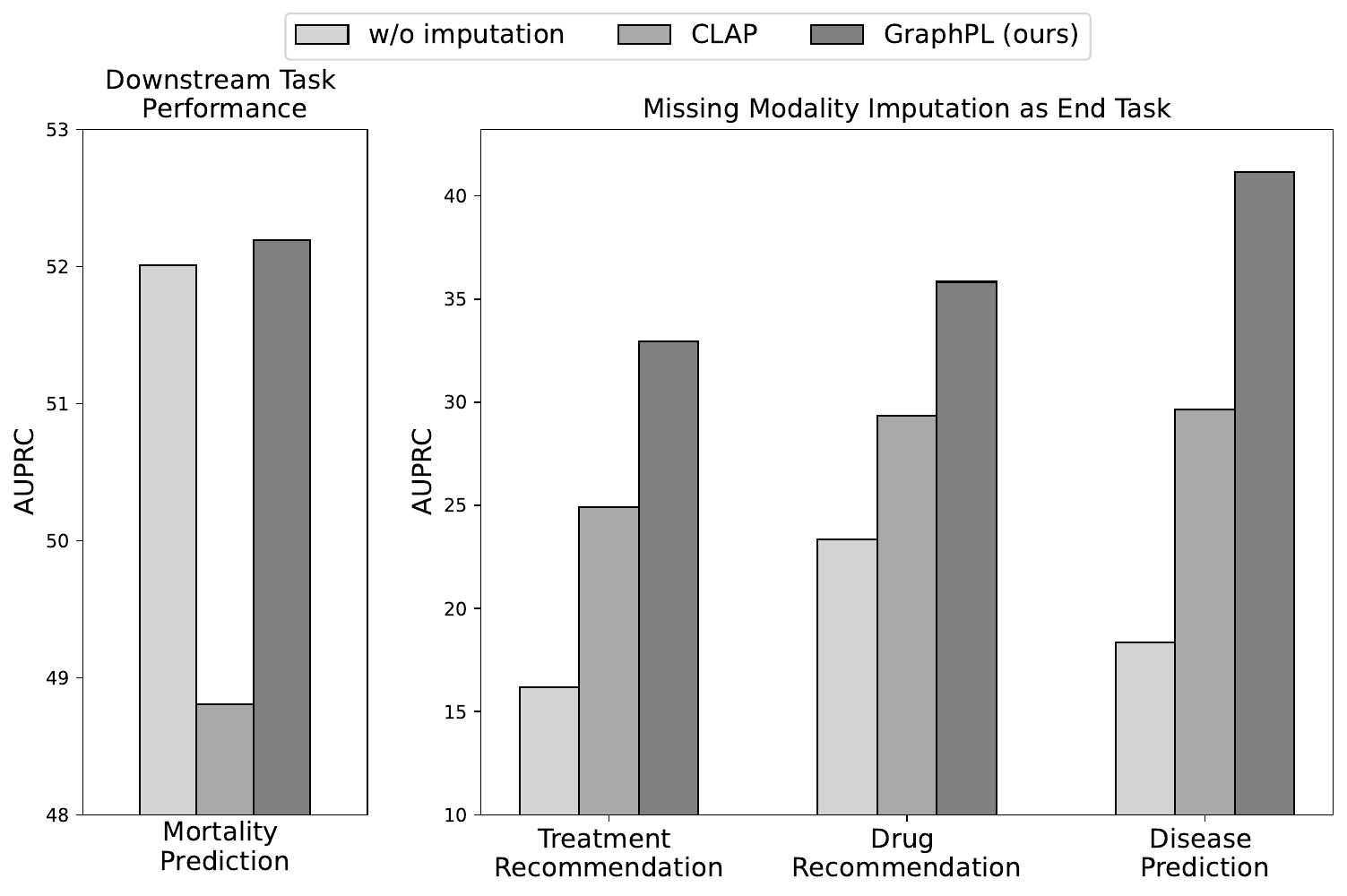}
   \caption{The performance of different methods on the real-world EHR dataset eICU~\cite{pollard2018eicu}. By incorporating imputation tasks and leveraging GNNs to mitigate the modality collapse issue identified in CLAP~\cite{cui2024clap}, GraphPL achieves superior performance in both tasks.}
   \label{fig: baselines}
\end{figure}

\begin{figure*}[t]
    \centering
    \begin{subfigure}[b]{0.32\textwidth}
        \centering
        \includegraphics[width=\textwidth]{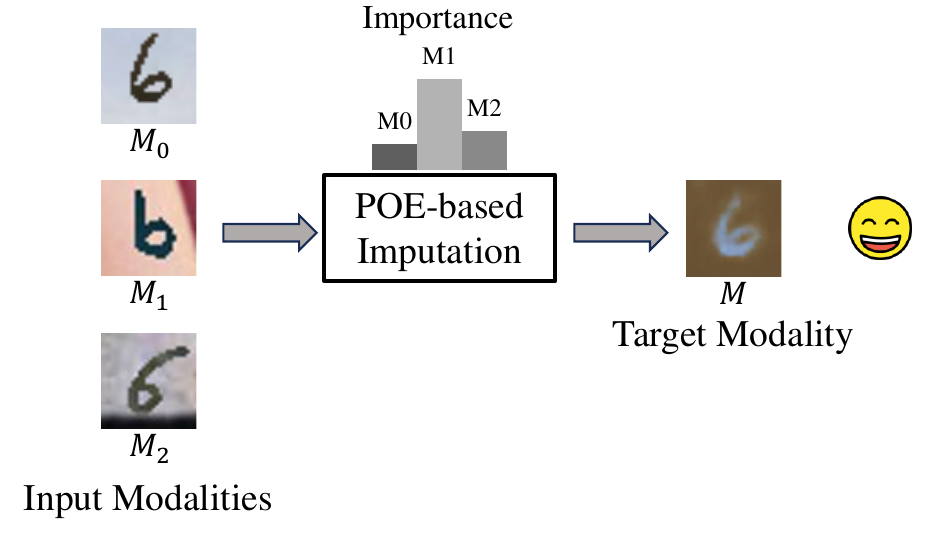}
    \end{subfigure}
    \begin{subfigure}[b]{0.32\textwidth}
        \centering
        \includegraphics[width=\textwidth]{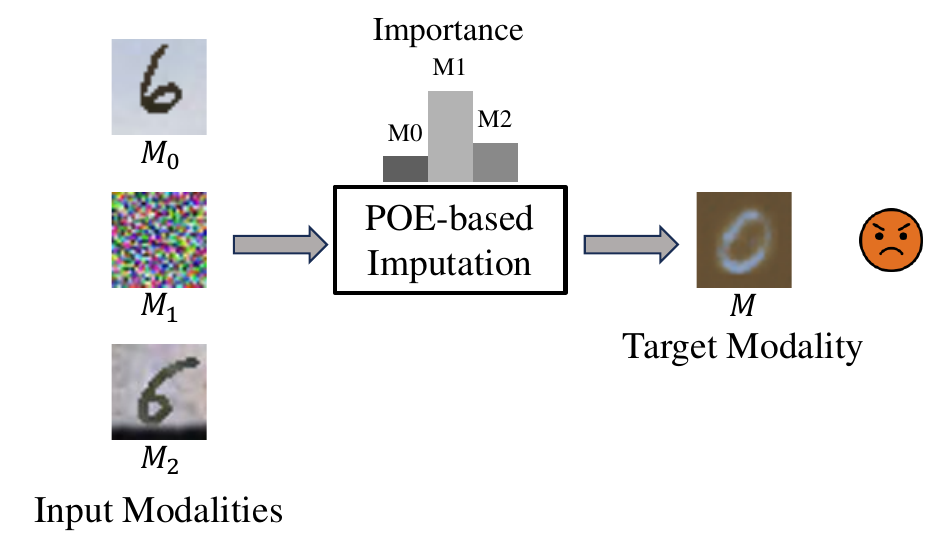}
    \end{subfigure}
    \begin{subfigure}[b]{0.32\textwidth}
        \centering
        \includegraphics[width=\textwidth]{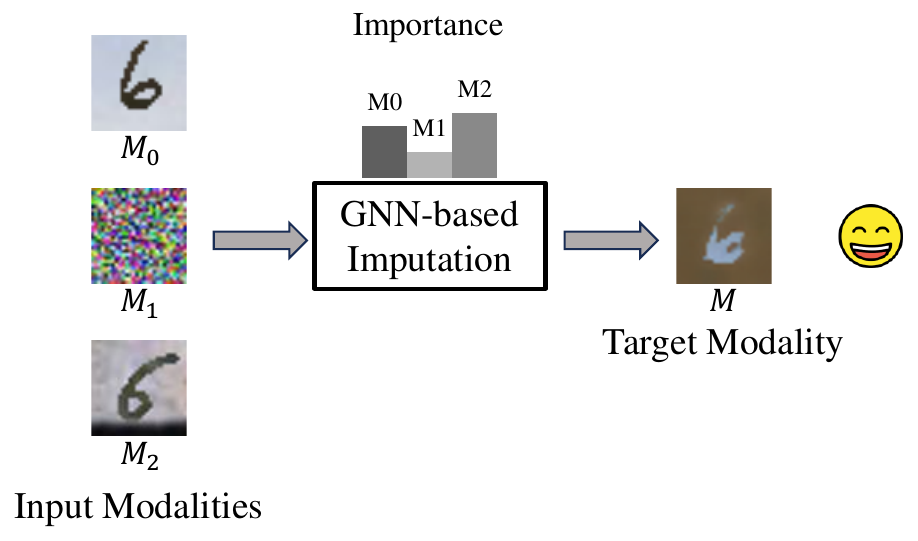}
    \end{subfigure}
    \caption{Existing POE-based method faced the \textbf{modality collapse issue}, due to its over-reliance on specific modality ($M_1$ here). In contrast, our GNN-based method merge the input information adaptively, exhibiting stronger robustness.}
    \label{fig: modality collapse}
\end{figure*}

Multi-modal learning~\cite{ngiam2011multimodal} extracting rich information from various modalities. Diverse modalities offer complementary analytical advantages via varied content, structure, and expression~\cite{baltruvsaitis2018multimodal}. It typically uses paired multi-modal data, with inter-modal correspondences per sample, helping models learn cross-modal associations for more comprehensive understanding.

However, in real-world scenarios, multiple modalities may be unavailable, and due to privacy concerns, data sharing is not possible~\cite{chen2022fedmsplit, yang2022cross}. Rajendran \etal~\cite{rajendran2023patchwork} introduce \textbf{patchwork learning} to address the situations where clients have incomplete and differing observed modalities while preserving data privacy. 
This scenario highlights the importance of missing modality imputation, which involves using the observed modalities to perform cross-generation of the missing modalities within the same data sample. As illustrated in Fig.~\ref{fig: baselines}, \textbf{imputation is essential for two primary reasons}: it enhances the quality of multi-modal representations by providing more comprehensive information, and it maintains data integrity, sometimes serving as an end task itself (e.g., imputing diagnostic information in healthcare)~\cite{ma2018kame, ma2017dipole}.

A limitation of existing methods~\cite{cui2024clap} is \textbf{modality collapse}~\cite{javaloy2022mitigating, wu2024muse}, where over-reliance on partial modality information leads to suboptimal performance and robustness, as shown in Fig.~\ref{fig: modality collapse}. These methods often use \ac{POE}~\cite{hinton2002poe} for fusion, which can exacerbate modality collapse~\cite{shi2019mmvae}.

To address the modality collapse issue, we propose \textbf{GraphPL}, a framework leveraging \ac{GNN} to dynamically fuse observed modalities. It constructs a modality-modality graph with nodes representing modalities and edges denoting interactions. A message-passing mechanism aggregates information across modalities, adapting to varying input modalities and noise.
Experiment results demonstrate that GraphPL is \textbf{effective}: it improves generation quality by \textbf{8.8\%, 13.9\%, and 4.8\%} on simulated benchmark datasets, and achieves gains of \textbf{11.5\%, 6.5\%, and 8.0\%} in disease diagnosis, drug recommendation, treatment recommendation, respectively, on the real-world EHR dataset eICU. Our code is available at: \url{https://github.com/LumionHXJ/GraphPL}. In summary, our key contributions are as follows:
\begin{itemize}[
  leftmargin=0.8em,
  itemsep=0em
]
    \item To address the modality collapse issue, we propose GraphPL, which leverages \ac{GNN} for modality fusion, allowing for flexible integration of input information. 
    \item  Experimental results demonstrate that GraphPL an average \textbf{9.2\%} improvement in imputation tasks across multiple simulated benchmark datasets as well as on the real-world distributed EHR dataset eICU. On eICU, it attains an average improvement of \textbf{8.7\%} over baseline methods in imputation tasks, confirming its practical effectiveness.
    \item GraphPL exhibits robustness under varying noise conditions, effectively mitigating modality collapse and enhancing overall model stability.
\end{itemize} 

\section{GraphPL}
\label{sec: graphpl}

\begin{figure}[t]
  \centering
   \includegraphics[width=\linewidth]{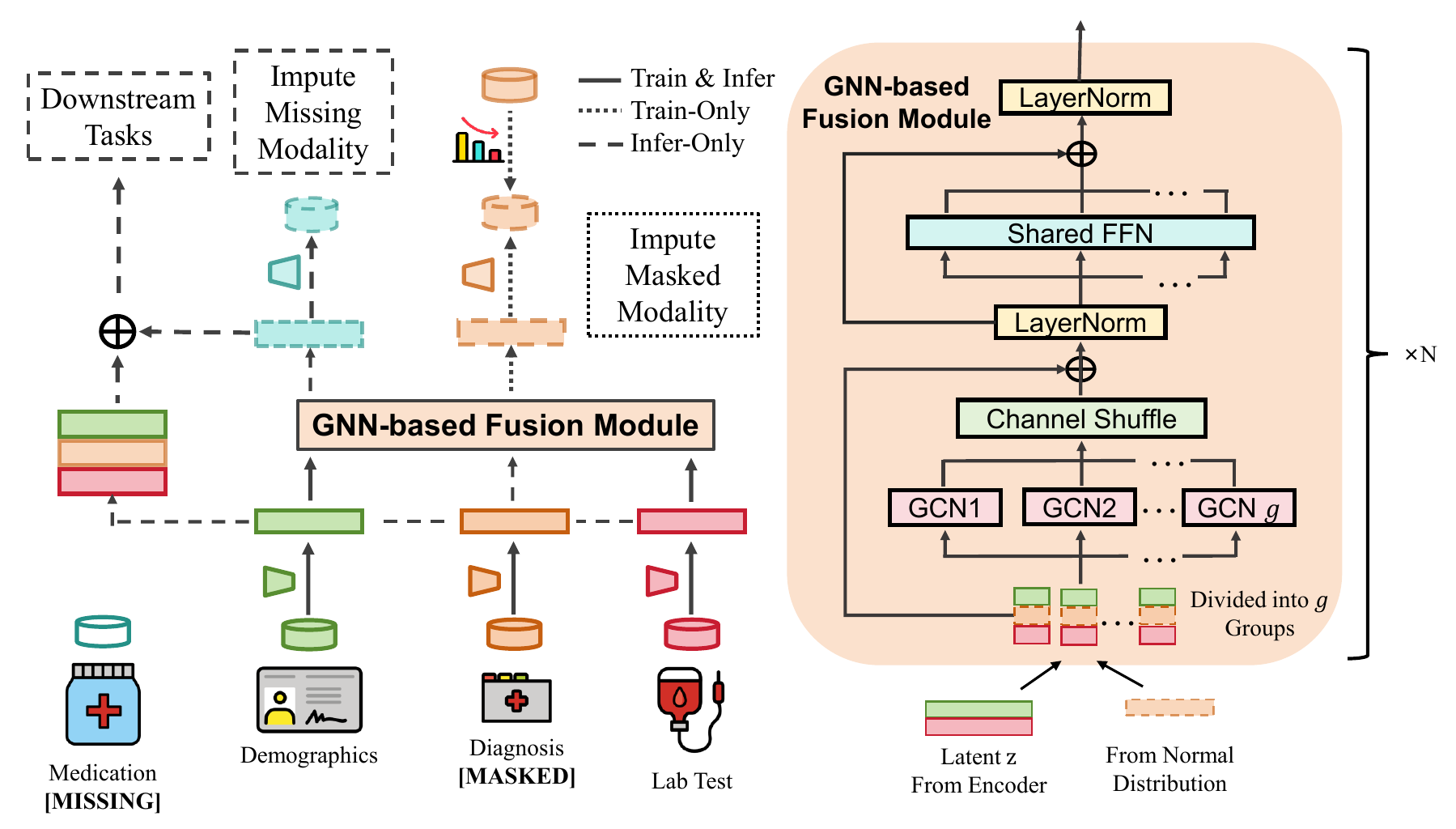}
   \caption{GraphPL’s overall pipeline (here, the client lacks the medication modality): During training, it masks one modality as target modality  and imputes it via other observed modalities as conditional ones. During inference, it uses all observed modalities as conditional modalities to predict the missing modality’s features for either imputation or downstream tasks.}
   \label{fig: pipeline}
\end{figure}

\begin{table*}[t]
\centering
\caption{Generation quality (\%) and representation quality (\%) on multiple benchmark datasets.}
\label{tab: benchmark}
\begin{tabular}{ccccccc}
\toprule
\multirow{2}{*}{Method} & \multicolumn{2}{c}{PolyMNIST}                           & \multicolumn{2}{c}{MST}                                 & \multicolumn{2}{c}{Quad-CelebA}                         \\
                        & \textbf{GQ} (\(\uparrow\)) & \textbf{RQ} (\(\uparrow\)) & \textbf{GQ} (\(\uparrow\)) & \textbf{RQ} (\(\uparrow\)) & \textbf{GQ} (\(\uparrow\)) & \textbf{RQ} (\(\uparrow\)) \\ \midrule
MVAE                    & $9.7_{\pm0.7}$             & $96.8_{\pm0.5}$            & $14.3_{\pm1.5}$            & $93.4_{\pm2.0}$            & $47.8_{\pm0.5}$            & $54.7_{\pm0.6}$            \\
MMVAE                   & $42.8_{\pm4.5}$            & $98.2_{\pm0.5}$            & $55.2_{\pm5.9}$            & $98.6_{\pm0.3}$            & $56.1_{\pm0.7}$            & $56.4_{\pm0.7}$            \\
MoPoE                   & $48.8_{\pm1.7}$            & $98.4_{\pm0.3}$            & $54.9_{\pm3.3}$            & $98.4_{\pm0.4}$            & $58.9_{\pm0.3}$            & $56.0_{\pm0.7}$            \\
CLAP                    & $46.8_{\pm3.1}$            & $96.9_{\pm0.7}$            & $51.7_{\pm6.7}$            & $98.7_{\pm0.4}$            & $60.1_{\pm0.6}$            & $57.3_{\pm0.1}$            \\
GraphPL                 & $\textbf{57.6}_{\pm5.0}$   & $\textbf{98.5}_{\pm0.2}$   & $\textbf{69.1}_{\pm6.3}$   & $\textbf{99.2}_{\pm0.4}$   & $\textbf{64.9}_{\pm0.7}$   & $\textbf{62.5}_{\pm0.6}$   \\ \bottomrule
\end{tabular}
\end{table*}

GraphPL is designed for the distributed multi-modal patchwork learning scenario, where clients have incomplete and diverse visible modalities, and data sharing is prohibited by privacy constraints.

The pipeline of GraphPL is illustrated in Fig.~\ref{fig: pipeline}. We implement an individual VAE~\cite{kingma2013vae} for each modality. During the forward pass, the input modalities are categorized into two groups: \textbf{conditional modalities} $x_{\text{cond}}$ (available as input) and \textbf{target modalities} $x_{\text{target}}$ (those to be imputed). The conditional modalities are first encoded into latent representations $z$ using their respective encoders. These features are then fused by a GNN-based fusion module to generate the features for target modalities. The resulting features can either be passed to decoders to reconstruct target modalities or concatenated with the features from conditional modalities for use in downstream tasks.

\subsection{GNN-based Fusion Module}
\label{sec: graph}

Previous methods~\cite{cui2024clap, wu2018mvae, sutter2021mopoe, shi2019mmvae} use \ac{POE} to fuse features of conditional modalities, the process can be overly influenced by partial distributions~\cite{shi2019mmvae}, facing the modality collapse issue. To overcome the limitations, we employ a \ac{GNN} for dynamic and adaptive modality fusion. Each conditional modality is represented as a node in a complete modality–modality graph, enabling full interaction across all available inputs. When imputing a target modality, it is introduced as a virtual node connected to all conditional modalities, allowing customized fusion for each target.

In the GNN fusion module, node embeddings for conditional modalities are derived from their VAE encoders, while target modalities are initialized via sampling from a standard normal distribution. Our fusion module consists of multiple identical modules, each composed of a grouped GCNConv~\cite{kipf2016gcn} followed by two-layer FFN. Channel shuffling~\cite{zhang2018shufflenet}  is applied after each GCNConv to promote cross-group information flow. LayerNorm stabilizes training and ensures output features remain compatible with the scale of encoder-derived features, facilitating effective modality imputation.

\subsection{Training} 

\textbf{Local Rounds.}
During local round on client $C^i$, GraphPL iteratively treats each observed modality as the target modality, with rest observed modalities serving as conditional modalities. Unlike methods optimizing the ELBO~\cite{kingma2013vae} of \( \log p(x_{\text{target}}) \), we directly model the conditional log-likelihood:
\vspace{-6pt}
\begin{equation}
        \log p(x_{\text{target}} | x_{\text{cond}}) =\mathbb{E}_{p(z|x_{\text{cond}})}  \log p(x_{\text{target}}|z). \\
    \label{eq: prob target}
\end{equation}

Since direct sampling \( z \sim p(z|x_{\text{cond}}) \) is infeasible, we introduce the single-modality reconstruction task and approximate it by VAE encoder output \( q(z|x_{\text{cond}})\approx p(z|x_{\text{cond}}) \)~\cite{kingma2013vae}. The imputation loss is thus defined as:
\vspace{-6pt}
\begin{equation}
    \mathcal L_{\text{impute}} = - \mathbb{E}_{q(z | x_{\text{cond}})} \log p(x_{\text{target}} | z).
    \label{eq: graph impute}
\end{equation}

Additionally, the single-modality reconstruction task is incorporated for each observed modality using its dedicated VAE, with loss similar to \(\beta\)-VAE~\cite{higgins2017betavae}. The total local loss combines both objectives:
\vspace{-6pt}
\begin{equation}
    \mathcal L_{\text{local}} = \mathcal L_{\text{impute}} + \lambda \mathcal L_{\text{single}},
\label{eq: total loss}
\end{equation}
with \(\lambda\) as a balancing hyperparameter.

\textbf{Global Rounds.}
After certain local rounds, each client uploads the encoder and decoder of its observed modalities, as well as the GNN parameters, to the server. The server then uses FedAvg~\cite{mcmahan2017fedavg} to update the global model parameters. 

\subsection{Inference}
\label{sec: infer}
During the inference phase, we serves two purposes: \textbf{imputing missing modalities and supporting downstream tasks}. For imputation, the missing modalities features output by the GNN module are passed to corresponding decoders trained by clients that possess the corresponding modality and shared via communication. For downstream tasks, the latent features of both observed and imputed modalities are concatenated to form an enriched representation for task-specific models.

\section{Experiments}
\label{sec: exp}

\subsection{Experiment Settings}
We evaluate GraphPL and baselines (MVAE~\cite{wu2018mvae}, MMVAE~\cite{shi2019mmvae}, MoPoE~\cite{sutter2021mopoe}, CLAP~\cite{cui2024clap}) on non-distributed benchmarks (PolyMNIST, MST, Quad-CelebA~\cite{sutter2021mopoe}) and real-world \ac{EHR} dataset eICU~\cite{pollard2018eicu}. The patchwork is constructed by randomly dropping some modalities on each client independently. Training uses FedAvg~\cite{mcmahan2017fedavg} for distributed learning and the Adam optimizer with batch size 256. Each experiment runs 5 times (same patchwork setup, different data splits), with results reported as means and STDs.

\textbf{Evaluation Metrics:} 1) \textbf{Generation Quality (GQ):} For missing modality imputation, pre-trained classifiers check if imputed class labels match ground truth. 2) \textbf{Representation Quality (RQ):} For downstream tasks, a logistic regression classifier is trained on each client using concatenated features; quality is measured via local test set classification accuracy.
    
\subsection{Experiments on Non-distributed Benchmarks}
\label{subsec: benchmarks}
Following \cite{cui2024clap}, we conduct experiments on same non-distributed benchmarks. Due to the number of modalities in the Bimodal CelebA dataset is limited, following Nair \etal~\cite{nair2023unite}, we extracted Canny edges from face images as sketches and generated face segmentation maps using FaRL \cite{zheng2021farl}, resulting in the \textbf{Quad-CelebA} dataset. To construct statistical heterogeneity for these non-distributed benchmarks, for PolyMNIST and MST, we perform class-imbalanced sampling to ensure that each client contains only some classes; for Quad-CelebA, we assign different proportions of male samples to each client. All of these experiments use 5 clients.

The results are shown in Table~\ref{tab: benchmark}. MVAE fails as it uses all observed modalities for training and cannot learn missing modality imputation. MoPoE performs better by using all observed modality combinations but incurs higher computational cost. Our method, GraphPL, achieves optimal performance due to its use of \ac{GNN} for flexible modality integration, demonstrating its effectiveness.

\subsection{Experiments on Distributed Dataset eICU}
\label{subsec: eicu}

\begin{table*}[t]
\centering
\caption{The experimental results on the eICU dataset, with 10, 20, and 50 hospitals participating, respectively. \textit{Treatment}, \textit{Medication}, and \textit{Diagnosis} represent the three missing modality imputation tasks of treatment recommendation, drug recommendation, and disease diagnosis, respectively. \textit{Mortality} represents the downstream task of mortality prediction.}
\begin{tabular}{cccccc}
\toprule
\multirow{2}{*}{\textbf{Hospitals}} & \multirow{2}{*}{\textbf{Method}} & \multicolumn{3}{c}{\textbf{GQ}}                                                 & \textbf{RQ}              \\
                                    &                                  & Treatment                & Medication               & Diagnosis                 & Mortality                \\ \hline
\multirow{5}{*}{10}                 & MVAE                             & $21.5_{\pm2.1}$          & $25.2_{\pm2.2}$          & $25.6_{\pm12.0}$          & $52.0_{\pm2.1}$          \\
                                    & MMVAE                            & $16.3_{\pm0.8}$          & $23.5_{\pm1.5}$          & $18.8_{\pm6.6}$           & $44.6_{\pm3.2}$          \\
                                    & MoPoE                            & $26.4_{\pm3.9}$          & $29.8_{\pm3.0}$          & $30.0_{\pm12.0}$          & $47.1_{\pm4.3}$          \\
                                    & CLAP                             & $24.9_{\pm3.7}$          & $29.4_{\pm2.3}$          & $29.6_{\pm11.9}$          & $48.8_{\pm4.3}$          \\
                                    & GraphPL                          & $\textbf{32.9}_{\pm8.1}$ & $\textbf{35.8}_{\pm2.5}$ & $\textbf{41.2}_{\pm14.3}$ & $\textbf{52.2}_{\pm3.5}$ \\ \hline
\multirow{5}{*}{20}                 & MVAE                             & $19.9_{\pm2.3}$          & $23.8_{\pm1.7}$          & $20.2_{\pm0.7}$           & $49.7_{\pm3.3}$          \\
                                    & MMVAE                            & $15.4_{\pm0.9}$          & $22.6_{\pm1.5}$          & $16.9_{\pm0.6}$           & $43.6_{\pm3.6}$          \\
                                    & MoPoE                            & $22.8_{\pm2.9}$          & $26.8_{\pm2.3}$          & $24.0_{\pm0.9}$           & $46.4_{\pm3.6}$          \\
                                    & CLAP                             & $22.5_{\pm2.6}$          & $26.6_{\pm2.1}$          & $24.8_{\pm0.9}$           & $47.4_{\pm3.3}$          \\
                                    & GraphPL                          & $\textbf{29.2}_{\pm3.9}$ & $\textbf{32.6}_{\pm2.4}$ & $\textbf{32.1}_{\pm1.3}$  & $\textbf{49.9}_{\pm4.0}$ \\ \hline
\multirow{5}{*}{50}                 & MVAE                             & $17.2_{\pm0.5}$          & $22.7_{\pm0.6}$          & $15.6_{\pm0.7}$           & $\textbf{45.9}_{\pm1.9}$ \\
                                    & MMVAE                            & $14.4_{\pm1.5}$          & $22.1_{\pm0.5}$          & $14.7_{\pm0.8}$           & $39.9_{\pm2.1}$          \\
                                    & MoPoE                            & $19.0_{\pm1.1}$          & $24.3_{\pm0.6}$          & $18.7_{\pm1.2}$           & $42.5_{\pm1.5}$          \\
                                    & CLAP                             & $21.9_{\pm1.6}$          & $25.2_{\pm0.6}$          & $20.6_{\pm1.1}$           & $43.9_{\pm1.3}$          \\
                                    & GraphPL                          & $\textbf{25.4}_{\pm2.6}$ & $\textbf{29.9}_{\pm0.9}$ & $\textbf{26.6}_{\pm1.9}$  & $45.8_{\pm2.1}$          \\ \bottomrule
\end{tabular}
\label{tab: eicu}
\end{table*}

To validate real-world performance, we use the distributed \ac{EHR} dataset eICU~\cite{pollard2018eicu}, with each hospital as a client. Data preprocessing follows MUSE~\cite{wu2024muse}, with input modalities: demographics, diagnosis (ICD-9 codes), treatment, medication (HICL codes), lab values, and vital signals. For diagnosis, treatment, and medication, many of whose categories are limited in data, we select top categories by frequency until cumulative probability exceeds 90\%, encoding their multi-label data as multi-hot vectors for input.

Given varying real-world modality missing probabilities and imputation value (e.g., demographics are mostly complete, making imputation unmeaningful), we construct patchworks by randomly dropping one modality from diagnosis, treatment, or medication per client. This lets the model handle three key imputation tasks: disease diagnosis, drug recommendation, and treatment recommendation.
Mortality prediction serves as the downstream task to evaluate representation quality. Both tasks use AUPRC due to class imbalance for assessment. 

Table~\ref{tab: eicu} presents the experimental results. GraphPL outperforms other baselines in both generation quality and representation quality across different numbers of participating hospitals, further confirming its advantages in real-world distributed scenarios.

\subsection{Addressing the Modality Collapse Issue}
\label{subsec: robust}

\begin{figure}[tb]
  \centering
  \includegraphics[width=\linewidth]{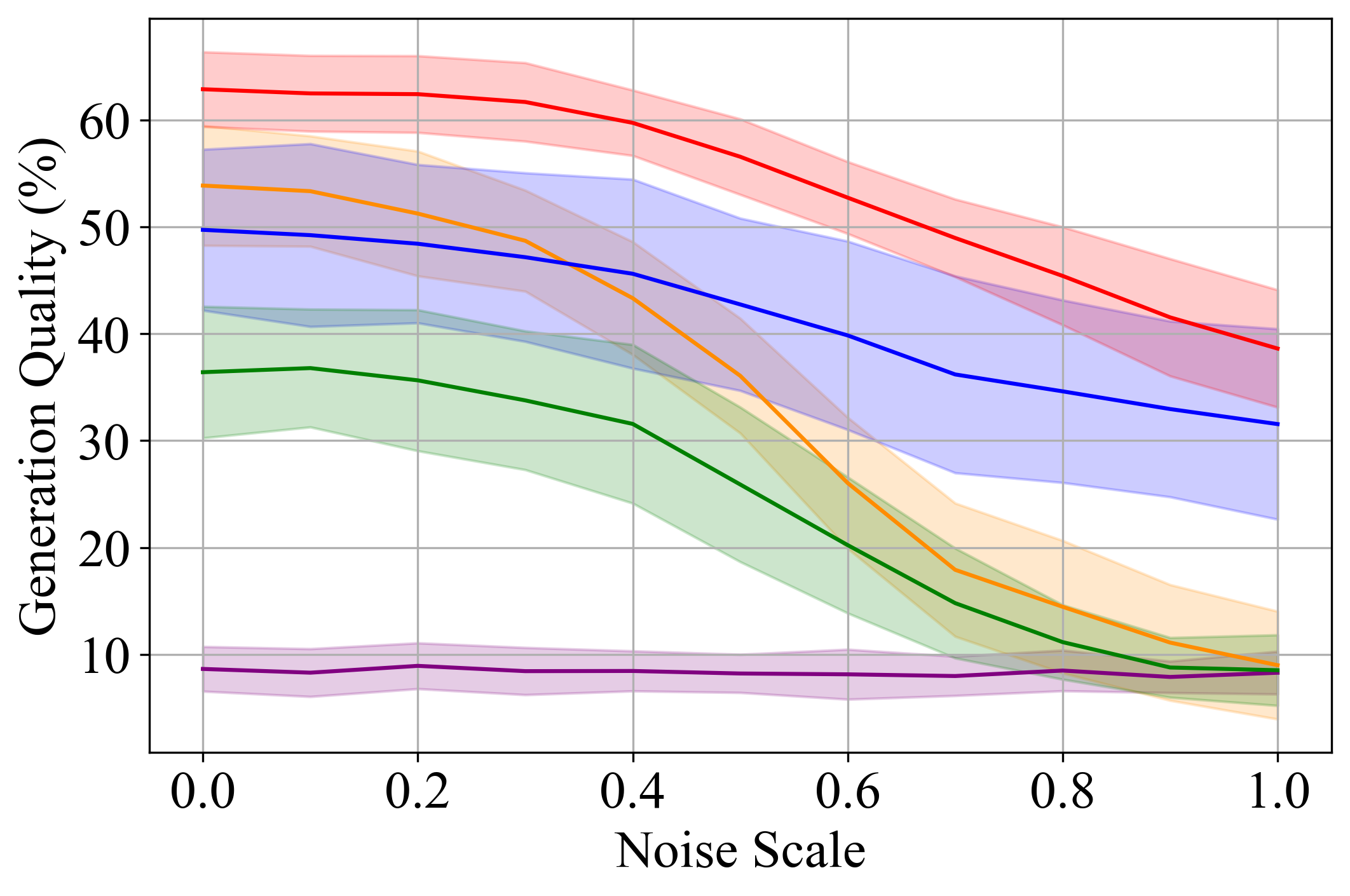}
  \caption{Generation quality of different methods (\textbf{\textcolor{mplpurple}{MVAE}, \textcolor{mplgreen}{MMVAE}, \textcolor{mplblue}{MoPoE}, \textcolor{darkorange}{CLAP}, \textcolor{mplred}{GraphPL}}) under varying noise scales.}
  \label{fig:grid}
\end{figure}

To verify that GraphPL alleviates the modality collapse issue, we conduct experiments on PolyMNIST by adding various noise (via interpolation between real and noisy images) to each observed modality during inference.
Fig.~\ref{fig:grid} shows generation quality changes with increasing noise scale when imputing a specific missing modality across methods. Metrics reflect the lowest generation quality under noise in different observed modalities.
Baselines using \ac{POE} for multi-modal feature fusion tend to rely on partial modalities, thus lacking robustness. In contrast, GraphPL uses \ac{GNN} to dynamically fuse all observed modalities, maintaining better generation quality at high noise and thus mitigating modality collapse.

\section{Conclusions}
In this paper, we explore the use of \ac{GNN} for multi-modal feature fusion in patchwork learning, achieving an effective and efficient patchwork learning method across various benchmark datasets. Furthermore, compared to existing methods, GraphPL is better at balancing the use of information from different modalities and alleviates the modality collapse issue. We hope this work will contribute to the development of the emerging field of patchwork learning.

\vfill\pagebreak

\section{Acknowledgement}
The work of Xingjian Hu, Jianhua Zhu and Liangcai Gao is supported by the projects of  Beijing Nova Interdisciplinary Program (20240484647) and National Natural Science Foundation of China (No. 62376012), which is also a research achievement of State Key Laboratory of Multimedia Information Processing, National Engineering Research Center of New Electronic Publishing Technologies and Key Laboratory of Science, Technology and Standard in Press Industry (Key Laboratory of Intelligent Press Media Technology).

\bibliographystyle{IEEEbib}
\bibliography{strings,refs}

\end{document}